This is a **PREPRINT**

"This page has been intentionally left blank."

# BN-HTRd: A Benchmark Dataset for Document Level Offline Bangla Handwritten Text Recognition (HTR) and Line Segmentation

**Md. Ataur Rahman, Nazifa Tabassum, Mitu Paul, Riya Pal**
*Premier University, Dept. of CSE, Chittagong, Bangladesh.*

**Mohammad Khairul Islam**
*University of Chittagong, Dept. of CSE, Chittagong, Bangladesh.*

CONTENTS

## 1.1 ABSTRACT

W E introduce a new *dataset*[1] for offline Handwritten Text Recognition (HTR) from images of Bangla scripts comprising words, lines, and document-level annotations. The `BN-HTRd` dataset is based on the BBC Bangla News corpus, meant to act as ground truth texts. These texts were subsequently used to generate the annotations that were filled out by people with their handwriting. Our dataset includes 788 images of handwritten pages produced by approximately 150 different writers. It can be adopted as a basis for various handwriting classification tasks such as end-to-end document recognition, word-spotting, word or line segmentation, and so on. We also propose a scheme to segment Bangla handwritten document images into corresponding lines in an unsupervised manner. Our line segmentation approach takes care of the variability involved in different writing styles, accurately segmenting complex handwritten text lines of curvilinear nature. Along with a bunch of pre-processing and morphological operations, both Hough line and circle transforms were employed to distinguish different linear components. In order to arrange those components into their corresponding lines, we followed an unsupervised clustering approach. The average success rate of our segmentation technique is 81.57% in terms of *FM* metrics (similar to $F$ measure) with a mean Average Precision ($mAP$) of 0.547.

## 1.2 INTRODUCTION

Data is the new oil in this era of the digital revolution. In order to make decisions through automatic and semi-automatic systems that employ machine learning (ML) and artificial intelligence (AI), we need to convert the handwritten documents in government, and non-government organization such as those in banks or that involves legal decision making. Although Bangla is one of the most highly spoken languages, so far, not too much attention has been given to the task of end-to-end handwritten text recognition from the document of Bangla scripts. Because of the lack of document-level (full page) handwritten datasets, we are unable to make use of the capabilities of modern ML algorithms in this domain.

This paper introduces the most extensive dataset named `BN-HTRd`, for Bangla handwritten images to support the advancement of end-to-end recognition of documents and texts. Our dataset contains a total of 788 full-page images collected from 150 different writers. With a staggering 1,08,147 instances of handwritten words, distributed over 13,867 lines and 23,115 unique words, this is currently the largest and most comprehensive dataset in this field (see Table 1.1 for complete statistics). We also provide the `lines` and ground truth annotations for both `full-text` and `words`, along with the segmented images and their positions. The contents of our dataset comes from a diverse news category (see Table 1.2), and annotators of different ages, genders and backgrounds, having variability in writing styles.

Segmenting document images into their most fundamental parts, such as words and text lines, is regarded as the most challenging problem in the domain of handwritten document image recognition, where the scripts are curvilinear in nature.

---

[1]`BN-HTRd` Dataset: https://data.mendeley.com/datasets/743k6dm543

Table 1.1 Statistics of the Dataset

| Types | Counts |
|---|---|
| Number of writers | 150 |
| Total number of images | 788 |
| Total number of lines | 13,867 |
| Total number of words | 1,08,147 |
| Total number of unique words | 23,115 |
| Total number of punctuation's | 7,446 |
| Total number of characters | 5,74,203 |

Table 1.2 Contents of Dataset According to News Categories

| Content Type | Documents | # of Pages |
|---|---|---|
| Sports | 41 | 202 |
| Coronavirus & Effected | 29 | 164 |
| Corona Treatment & Vaccine | 16 | 90 |
| Election | 17 | 88 |
| Story of a Lifetime | 09 | 67 |
| History | 06 | 34 |
| Political | 04 | 24 |
| Mission of Space | 04 | 16 |
| Corruption | 04 | 13 |
| Economy | 04 | 10 |
| Others | 17 | 80 |

Thus, we also present an unsupervised `segmentation` methodology of handwritten documents into corresponding `lines` along with our dataset. The proposed approach's main novelties consist of extending and combining some of the earlier reported works in the field of text line segmentation.

## 1.3 RELATED WORK

The task of handwriting recognition captivated researchers for nearly a half-century. Although such initial triumph began with simple handwritten digit recognition, the first-ever massive character-level recognition task was arranged in 1992 by the `First Census of Optical Character Recognition System Conference` [1]. After that, researcher slowly started building sentence-level [2] as well as document-level [3] offline handwritten datasets for English. This dataset (`IAM`) was subsequently used to initiate one of the most popular handwriting recognition shared tasks - `ICDAR` [4].

We found only a handful of Bangla handwriting datasets, among which the majority of those are isolated character datasets. One such dataset is the `BanglaLekha-Isolated` [5], which is comprised of a set of 10 numerals, 50 basic characters, and 24 carefully curated compound characters. For each of the 84 character samples, they accumulated 2000 individual images. The resulting dataset in-corporates a total of 1,66,105 images of handwritten characters after discarding the scribbles. It also holds information regarding the age and gender of the subjects from whom the writing samples were obtained.

Another multipurpose handwritten character dataset named `Ekush` [6] consists of 3,67,018 characters. This dataset was collected from different regions of Bangladesh with equal numbers of male and female writers and varying age groups. The dataset contains a collection of modifiers as well, which is missing from other similar character-level datasets. Apart from this, the `ISI` [7] and `CMATERdb` [8] datasets are two of the oldest character based handwritten dataset for the Bangla language.

The only dataset that resembles our own dataset in terms of word-level annotation is the `BanglaWriting` [9] dataset. It includes the handwriting of 260 people of diverse ages and personalities. The authors used an annotation tool to annotate the pages with bounding boxes containing the words Unicode representation. This dataset comprises a total of 32,787 characters and 21,234 words, having a vocabulary size of 5,470. Although all of the bounding boxes of word labels were manually produced, they did not provide the actual ground truth of those pages from where they generated the writings. On top of that, almost all of the pages were short in length and can be comprehended as more like a paragraph instead of a full page document.

Proper segmentation of text lines and words from document images containing handwritings is an essential task before any kind of recognition, such as layout analysis and authorship identification, etc. It is still deemed as a challenging task due to the (i) irregular spacing between words and (ii) variations of writing habits among different authors. The Hidden Markov Models (HMMs) [10] was extensively used in continuous speech recognition and single word handwriting recognition before the popularity of Neural based models. As the parameter estimation of HMM is more general, it leads to better recognition results initially, even with fewer pre-processing operations that mainly dealt with positioning and scaling [11].

Other initial approaches to text line segmentation were mainly based on connected component (CC) analysis [12, 13]. In such scenarios, the connected components' average width and height are first estimated using some form of Ad-hoc or statistical methods. Different lines were separated by Hough Line transformation, and some form of clustering scheme allowed to distinguish between each component that falls under distinct lines [14].

A local region-based text-line segmentation algorithm was proposed in [15]. The lines are simply segmented by using a horizontal projection-based approach. Text regions are detected locally considering the corresponding approximated skew angles, and also, considering skews of blocks, the proposed method outperformed all other approaches during that time.

In the Bangla script, a word can be horizontally partitioned into three adjoining zones - the lower zone, middle zone, and upper zone. The author of this article [16] used these zones in order to distinguish among different words within a line. Unfortunately, they only detected the words, which doesn't mean much without their location in the corresponding lines. Our approach relates that work in a way that the black pixels on the '*matra*' (contiguous upper zone) are automatically identified as segment points.

A more advanced work in this domain for segmenting the text-line/word images into sub-words using a graph modeling-based approach was done in [17]. They have achieved the sub-word-level segmentation while also considering the issue of displacement of the diacritics.

The use of convolutional neural networks with a combination of LSTM and other deep learning frameworks to detect and recognise the lines or words in an image became popular after 2017 [18, 19, 20, 21]. The results obtained are promising, which encouraged us to do more research in this direction. This, however, requires a lot of training data. Our dataset is mainly targeted to achieve this in our future endeavors.

## 1.4 DATA ANNOTATION

Annotation is a means of populating a corpus by examining something in the world and then recording the observed characteristics. The dataset is essentially organised into a particular model that helps to process the needed information. In this section, we will brief about the different stages of our dataset creation and annotation process.

### 1.4.1 Data Collection and the Source

As a first step, we have collected individual text documents from `BBC Bangla News`[2] as our ground truth data by automatically Crawling/Scraping the website. We mainly prefered this source for our dataset because the BBC Bangla News does not require any restrictions and has an open access policy[3] to their data for the general public. In most cases, we downloaded both the `TEXT` and `PDF` file for a particular news. Secondly, those pdf files and texts have been renamed according to a sequence (1 to 150) and placed in a separate folder (Fig. 1.1) before distributing them to different writers.

### 1.4.2 Data Distribution

For Data annotation, we have to distribute data among individuals. So, we provided the folders (having the text and pdf) to people of different ages and professions. To be specific, 85 of those data were given to the undergraduate students and the rest (65) to writers of different background. We also provided them with a sample (an annotated folder) and annotation instructions. In return, they wrote the contents of the file and gave us back the images or hard copies of the handwriting pages. Note that a single writer had to write multiple pages (up to 20) in most case because of the length of the news. Fig. 1.1 below represents the arrangements of our dataset for a single folder, and Fig. 1.2 shows a handwritten sample page.

### 1.4.3 Annotation Guidelines

As the initial handwriting's were gathered from 150 native writers, each individual was instructed before writing the provided text and ensured that they write the text naturally.

---

[2]https://www.bbc.com/bengali
[3]https://www.bbc.com/bengali/institutional-37289190

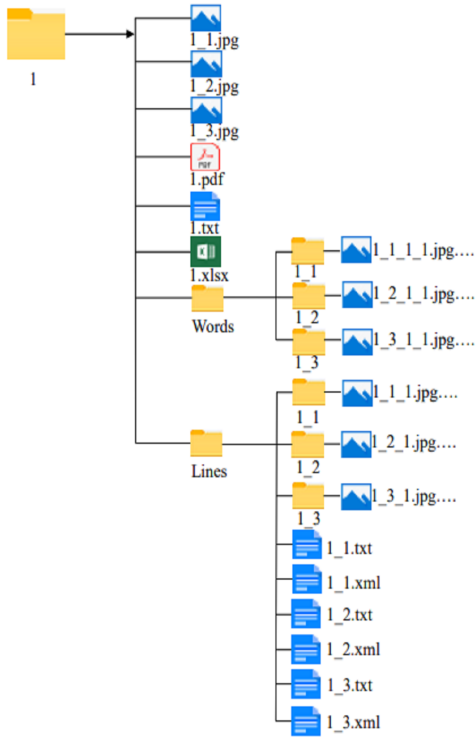
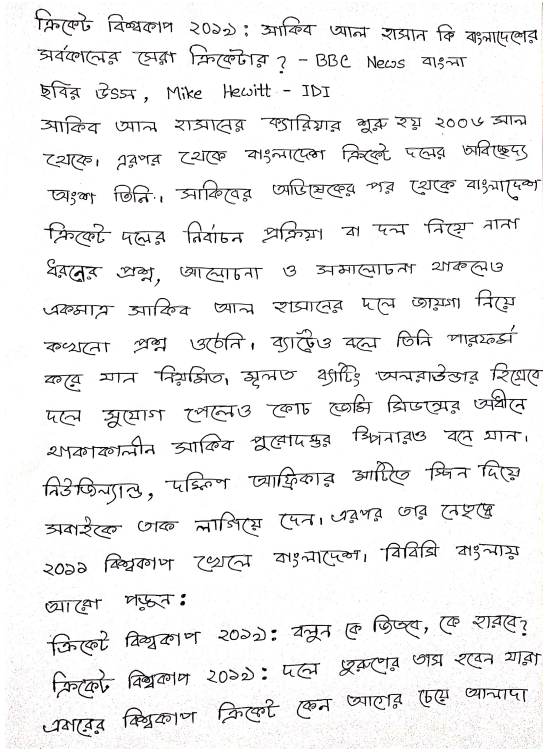

Figure 1.1  Dataset's Folder Structure    Figure 1.2  A Sample Image from the Dataset

Thus, we have provided the following guidelines to everyone:

1. Pages must be of A4 size and cannot be written on both pages.
2. If any line in the text file is missed while writing, it should be added at the end of the last page. Then we changed the ground truth text accordingly.
3. If the spelling of some word is wrong while writing, he/she should cut it with one pull, and the corrected one should be written next to it.
4. There is no need to end or start the line as it is in the text file. The following line can be started from where the previous one ends.
5. While scanning the written text (page) with the CamScanner app, the camera's resolution should be good so that the image is not blurred.

### 1.4.4  Annotation Scheme and Agreement

We collected and digitized the handwritten pages into images and performed skew correction. After that, we segmented the pages into images of line, individual words (digits, punctuation's) and created an Excel file for corresponding words. `Line` and `Word` folders are different for each image.

- Images are labeled by: `FolderNumber_PageNumber`.

- Lines are labeled by: `FolderNumber_PageNumber_LineNumber`.
- Words are labeled by: `FolderNumber_PageNumber_LineNumber_WordNumber`.
- Excel files are created including segmented words `Unicode` representation with their labeled id.

For the annotation of words, we have done it in two folds. We instructed the 85 students to crop the words from the scanned pages that they have written. The cropped images correspond to each of the words and are placed into separate folders having their page and line number. We followed this specific pattern of naming the words/lines so that their filename corresponds to their position in the images (see above naming convention). Students also filled out a `EXCEL` file with the corresponding word identifier (file name in the aforementioned format) and their *Unicode Text* (Bangla word). The rest of the 65 authors word annotation was done by us in a similar way.

For the line annotation, we have used a tool named `LabelImg`[4] in order to get YOLO and PascalVOC formatted[5] annotations. Note that students partially did only the word segment part as an assignment. We did the line segmentation and all the other task related to the dataset curation. From those line annotation, we programmatically extracted the corresponding lines and arranged them in subfolders. We have done this in order to apply deep learning frameworks such as YOLO/TensorFlow for line detection in a supervised way in our future research.

### 1.4.5  Data Correction

We made a video presentation for students to introduce the whole process of word annotation. After understanding the annotation process, students finally submitted their work, and we compared it word-by-word. Although their submission wasn't 100% accurate all the time, we corrected those faults manually. While examining, we tried to do it as much accurately as possible by following some rules such as:

1. The submitted images should match the given text file (line by line).
2. The cropped words (digits, punctuation's) have been checked individually.
3. The Id and Word column of the Excel file have been manually verified.
4. The sequence of cropped words have been compared against the Excel file.

## 1.5  LINE SEGMENTATION: METHODOLOGY

In this section, we describe our text line segmentation approach in details. Fig. 1.3 illustrates the process of `BN-HTRd` dataset collection and preparation (left) and the overall system architecture of the line segmentation pipeline (right).

---

[4]LabelImg: https://tzutalin.github.io/labelImg/
[5]YOLO/PascalVOC Format: https://cutt.ly/LvrTrCH

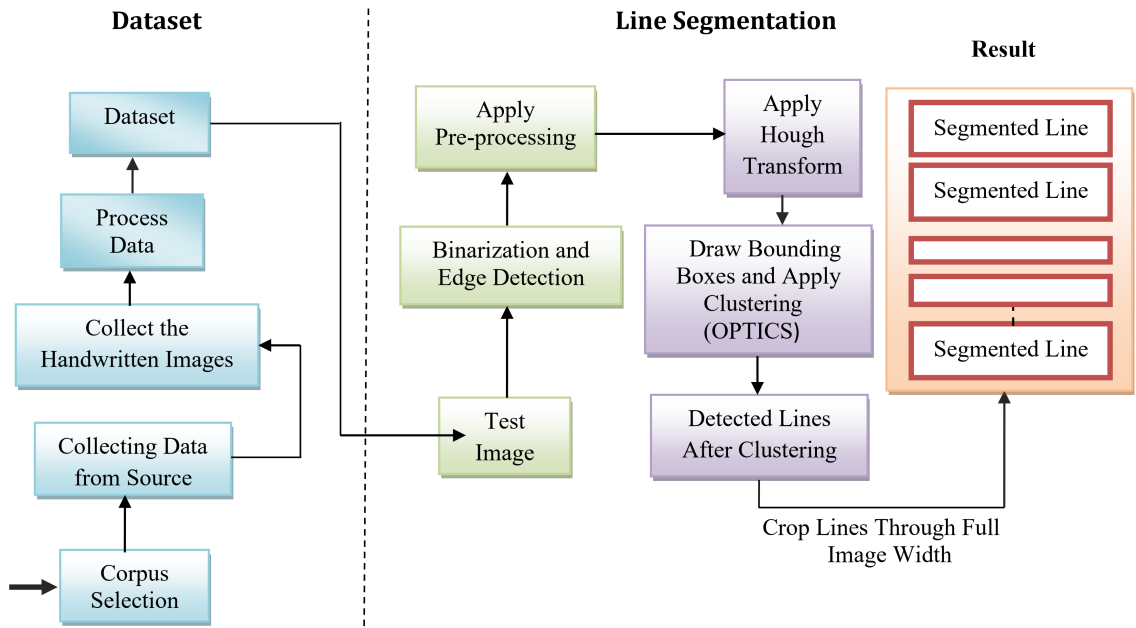

Figure 1.3  Overall System Architecture

### 1.5.1  Thresholding and Edge Detection

Thresholding or image binarization is a non-linear process that transforms a grayscale (or coloured) image to a binary image having only two levels (0 or 1) for representing each pixel considering the specified threshold value. In other words, if the pixel value is higher than the threshold, it is assigned one value (i.e., white), else it is assigned another value (i.e., black). We have incorporated a local OTSU's technique for this. Fig. 1.4 shows the result of the binarization process over an original image segment.

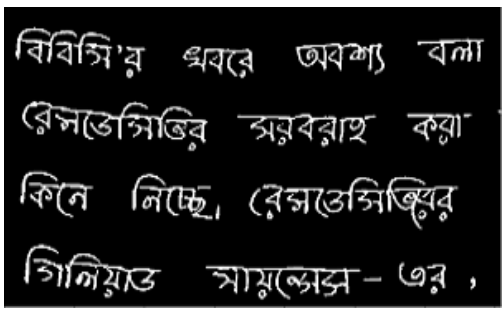 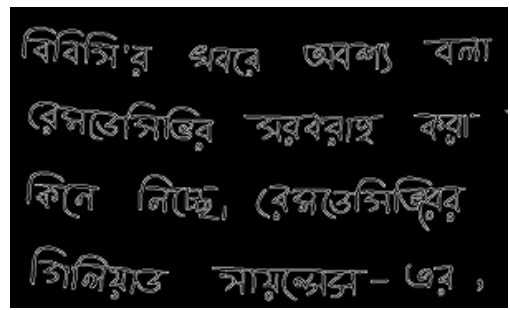

Figure 1.4  Thresholded Image     Figure 1.5  Canny Edge Detection

Edge detection is a procedure that extracts or highlights useful regions in an image having different objects and subsequently reduces the amount of non-useful pixels to be considered. We have used the Canny edge detection technique after binarization to highlight the edges of the handwritings (Fig. 1.5).

### 1.5.2 Morphological Operation and Noise Removal

In order to remove the small salt and paper type noise, we have used morphological opening followed by dilation to separate the sure foreground (Fig. 1.6) noise from the background. To find the sure foreground objects, we have used the distance transformation and subtracted it from the background. The resultant image of Fig. 1.7 shows the situation after these preprocessing steps.

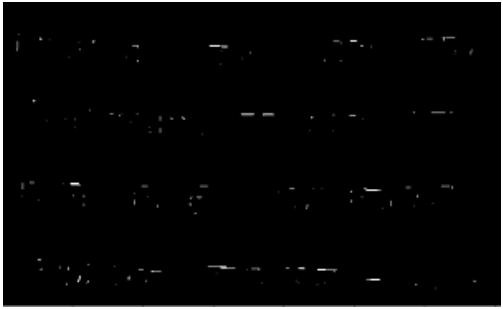
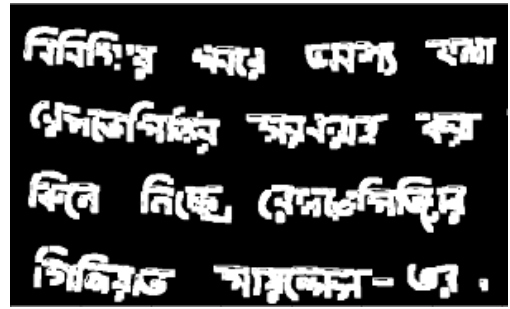

Figure 1.6  Sure Foreground (Noise)   Figure 1.7  After Removing Noise

### 1.5.3 Hough Line Detection

We used the `Hough Line Transform` to distinguish the horizontal continious lines ('matra') over the words and dialated them in order to thicken those lines so that each word acts as a connected component and separate words can be distinguished within a line. This will help us later on to draw the bounding box more accurately over the words. Fig. 1.8 shows the result.

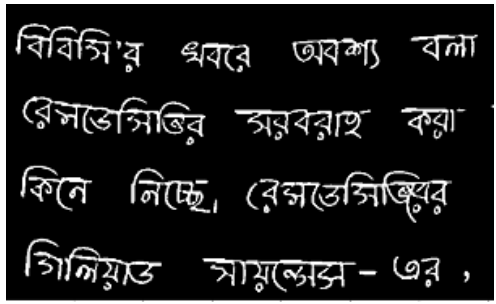
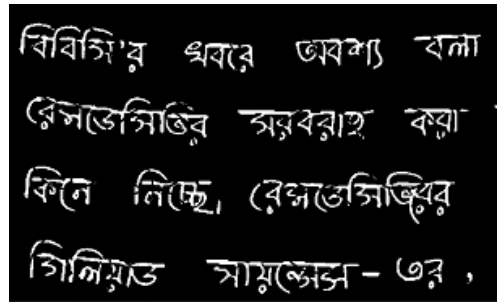

Figure 1.8  Hough Line Detection   Figure 1.9  Hough Circle Removal

### 1.5.4 Hough Circle Removal

Most of the time, two or more lines in a text document overlaps due to the circle like shape in Bangla scripts. We have used `Hough Circle Transform` to detect those circular object and break them apart so that two consecutive horizontal line doesn't form a connected component due to overlapping word segments. Fig. 1.9 illustrates the resultant partial image after hough circle removal operation.

### 1.5.5 Bounding Box

We used the `Connected Component (CC)` analysis to draw bounding boxes over each connected regions (Fig. 1.10). After that, we took boxes with a certain minimum area and determined the centre of the boxes (see the red dot inside the box) in order to use them in the next step for clustering.

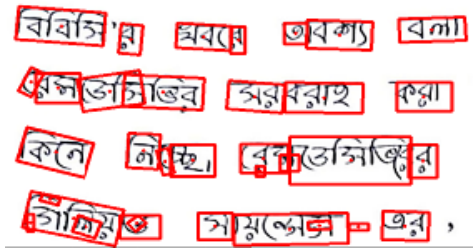
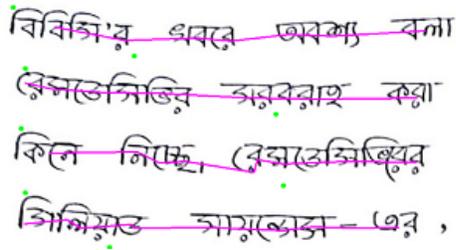

Figure 1.10  Bounding Box and Midpoints over the Connected Components

Figure 1.11  Add the Midpoints and Marking Highest and Lowest Points of Line

### 1.5.6 OPTICS Clustering

OPTICS Clustering stands for Ordering Points To Identify Cluster Structure. We used this algorithm over the `Y` co-ordinate (vertical-axis) of the midpoints that we found in the previous step (section 1.5.5). It mostly gives us the points that fall in a line within a single cluster. Thus each cluster after this operation represents separate text lines in the image. After performing the clustering, we have added the midpoints and got the line annotated Fig. 1.11. Note, however, that sometimes it fails to determine lines in the cases when lines are too close to each other, or the midpoints of the bounding box are determined incorrectly shows in Fig. 1.12. One of the advantages of using OPTICS clustering is that, unlike K-Means clustering, we do not need to provide the number of clusters (k) beforehand. Here it serves our purpose as the number of lines per document image is not fixed and depends on the authors writing style and page size.

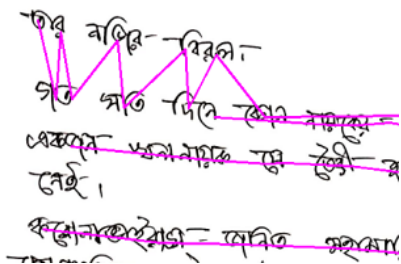
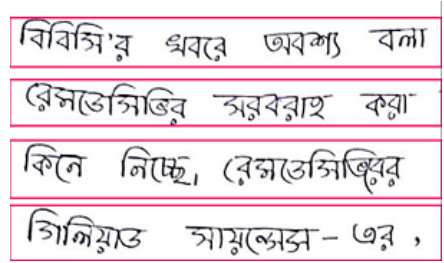

Figure 1.12  Failed to Cluster Properly

Figure 1.13  Bounding Box over the Lines

### 1.5.7 Line Extraction and Cropping

In order to visualise the lines, we have used a rectangle over each of them (Fig. 1.13). We crop individual lines taking the full length of the bounding box. We considered the bounding boxes' top and bottom points (see green dots in Fig. 1.11) from the connected components (section 1.5.5) to determine the height of the cropped lines. Fig. 1.14 delineates two of the line cropped through our method.

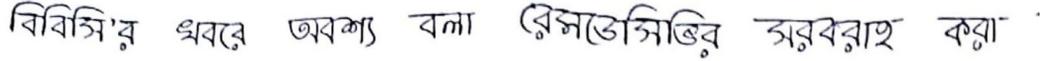

Figure 1.14  Two Cropped Lines from Fig 1.13

## 1.6  RESULTS AND EVALUATION

In this section, we will talk about our line segmentation's result. Fig. 1.15 shows the output lines detected through our method for different handwritten images.

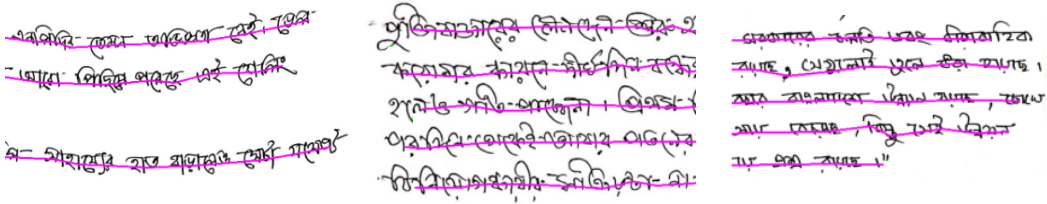

Figure 1.15  Output Lines for Different Types of Handwritings

### 1.6.1  Evaluation Metrics

Two bounding boxes (lines) are considered as a `one-to-one match` if the total matching pixels is greater than or equals to the evaluator's approved threshold ($T_a$). Let $N$ be the number of ground-truth elements, $M$ be the count of detected components, and $o2o$ be the number of one-to-one matches between $N$ and $M$, the `Detection Rate (DR)` and `Recognition Accuracy (RA)` are defined as follows:

$$DR = \frac{o2o}{N}, \quad RA = \frac{o2o}{M} \tag{1.1}$$

By combining the detection rate (DR) and recognition accuracy (RA), we can get the final performance metric $FM$ (similar to $F$ measure) using the equation below:

$$FM = \frac{2DR * RA}{DR + RA} \tag{1.2}$$

`Average Precision (AP)` on the other hand calculates the average values of Precision (P) for the corresponding Recall (R) over 0 to 1 with a interval of 0.1:

$$AP = \frac{1}{11} \sum_{r \in 0, 0.1, 0.2, ...1.0} P(r) \tag{1.3}$$

The `mean Average Precision (mAP)` is computed by using the mean AP across every class having a threshold (equivalent to $T_a$):

$$mAP = \frac{1}{n}\sum_{k=1}^{k=n} AP_k \qquad (1.4)$$

where $AP_k$ is the average precision of class $k$, and $n$ is the number of total classes.

### 1.6.2 Line Segmentation Results

We evaluated the performance of our algorithm for text line segmentation using equations 1.1–1.2 over a portion of our dataset[6] (150 images). The acceptance threshold used was $T_a = 80\%$. That is, if the bounding box of ground truth line and our detected line have an 80% match in terms of the pixel area, we considered it accurate. Here, we only calculated the result for 150 unique images from 788 images (one image from every 150 folders). The total number of lines from 150 images are 2915, and by applying our method, we got 3437 lines, from which 2591 lines match with the ground truth having the aforementioned threshold. So, The value of $N$ (ground truth) is 2915, value of $o2o$ is 2591 and $M$ is 3437. Now, using equation 1.1 and 1.2 we get:

$$DR = \frac{2591}{2915}, \quad RA = \frac{2591}{3437}, \quad FM = \mathbf{0.8157}$$

Apart from our own `BN-HTRd` dataset, we have also tried the ICDAR2013 dataset[7] containing 50 Bangla test images for the Handwriting Segmentation Contest [12]. The results obtained from our unsupervised approach that we have used to perform line segmentation for both datasets are presented in Table 1.3.

Table 1.3 Detailed Results on Two Datasets for Line Segmentation Based on FM (F Score)

| Evaluation Metrics | Datasets | |
|---|---|---|
| | BN-HTRd | ICDAR2013 |
| # of Images | 150 | 50 |
| N | 2915 | 872 |
| M | 3437 | 943 |
| o2o | 2591 | 695 |
| DR(%) | 88.88% | 79.7% |
| RA(%) | 75.38% | 73.7% |
| FM(%) | **81.57%** | 76.58% |

Table 1.4 Recall and Precision Values (11 point measurements)

| No. | Recall | Precision |
|---|---|---|
| 1. | 1.0 | 1.0 |
| 2. | 0.9 | 0.76 |
| 3. | 0.8 | 0.73 |
| 4. | 0.7 | 0.76 |
| 5. | 0.6 | 0.71 |
| 6. | 0.5 | 0.4 |
| 7. | 0.4 | 0.42 |
| 8. | 0.3 | 0.68 |
| 9. | 0.2 | 0.3 |
| 10. | 0.1 | 0.26 |
| 11. | 0.0 | 0.0 |

---

[6] Line Segmentation Results: `https://cutt.ly/cczzQ9i`
[7] ICDAR2013 Dataset: `https://cutt.ly/yvi8OrF`

From the results, we can see that the `ICDAR2013` dataset contains fewer images having less lines as compared to our own dataset, thus M and o2o varies. Hence, there is a difference in terms of performance of our algorithm for these two datasets, which is nearly 5% (81.57% Vs. 76.58%) in this case. Another reason is that, the images in the `ICDAR2013` dataset are smaller in terms of resolution, and this caused an intricacy as we have considered a standard resolution (width of at least 1000 pixels) while we run our system.

To have a more accurate idea of the performance of our approach, we further calculated the recall and precision values for each of the 150 images that we tested from our `BN-HTRd` dataset. We then took the highest precision values (Table. 1.4) for the recall values ranging $0.0 - 1.0$ and having an interval of 1.0. Fig. 1.16 depicts this scenario in terms of Recall Vs. Precision graph. Here we only took 11 values since we only need these values for calculating AP and mAP.

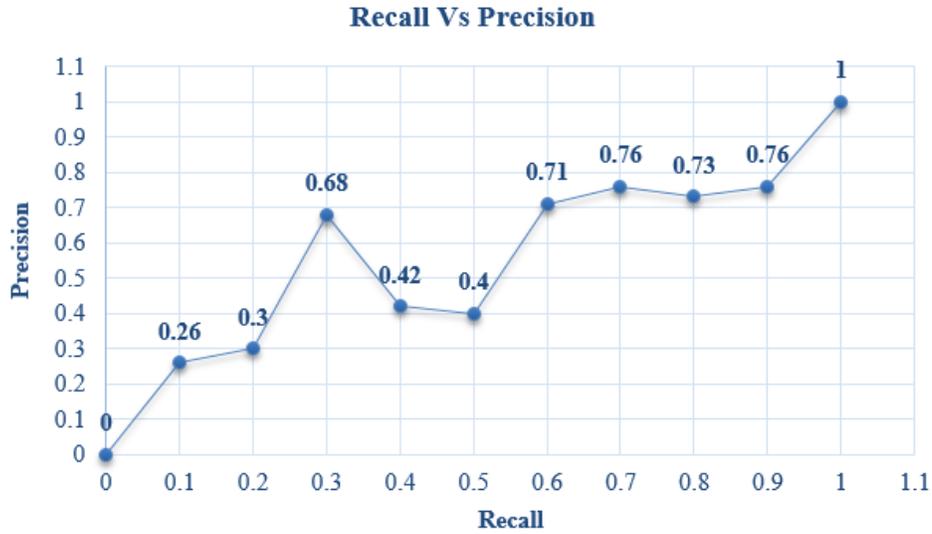

Figure 1.16  Recall v/s Precision Graph

Using equation 1.3 and the values from Table. 1.4, we get:

$$AP = \frac{1}{11}(1.0 + 0.76 + 0.73 + 0.76 + 0.71 + 0.4 + 0.42 + 0.68 + 0.3 + 0.26 + 0)$$

Thus the `average precision` for the `BN-HTRd` dataset is:

$$AP = \frac{1}{11}(6.02) = \mathbf{0.547}$$

As we performed only the line segmentation, here we have only one class ($n = 1$ in equation 1.4). So, In this case, AP and mAP remain the same. Thus, our final `mAP` for line segmentation is **0.547**.

## 1.7 CONCLUSION AND FUTURE WORK

Our endeavour in this paper was to lay the groundwork for future research on Bangla Handwritten Text Recognition (HTR). Keeping this in mind, we have collected and developed the largest ever dataset in this domain, having both text line and word annotations as well as the ground truth texts for full-page handwritten document images. We also propose a framework for segmenting the lines from the input documents. Initially, the input images are resized and converted into binarized frames. Then the image noises and shaded effects are removed. After that, a connected component based segmentation method is applied to segment the components (mostly words) in the image. We employed the OPTICS clustering on the bounding boxes from those segments in order to produce the final line segmentation. Our framework was able to achieve 81.57% in terms of FM score for line segmentation, having a mean average precision of 0.547 (mAP@0.8). In Bangla literature, many handwritten documents need to be converted into electronic version. Handwritten line segmentation is an important part towards that goal. For that purpose, this work will help push the research work on this determination one step forward. We aim to extend this work by incorporating word segmentation from the lines and recognizing individual words using deep learning models in future. Our dataset is ready to deal with these objectives. We look forward to the research community around the globe who will use this dataset to achieve:

> "`End-to-End Bangla Handwritten Image Recognition`"